\documentclass[11pt]{article}

\usepackage[preprint]{neurips_2020}

\usepackage[utf8]{inputenc} 
\usepackage[T1]{fontenc}    
\usepackage{hyperref}       
\usepackage{url}            
\usepackage{booktabs}       
\usepackage{amsfonts}       
\usepackage{nicefrac}       
\usepackage{microtype}      
\usepackage{graphicx}
\usepackage{subfigure}
\usepackage{wrapfig}
\usepackage{amsmath}

\PassOptionsToPackage{x11names}{xcolor}
\usepackage{xcolor}
\PassOptionsToPackage{algo2e,ruled}{algorithm2e}
\usepackage{algorithm2e}
\title{Biologically Inspired Mechanisms for Adversarial Robustness}

\author{%
  Manish V.~Reddy \\
  Institute for Applied Computational Science\\
  Harvard University\\
  Cambridge, MA 02138 \\
  \texttt{mvuyyuru@g.harvard.edu} \\
   \And
   Andrzej Banburski \\
   Center for Brains, Minds and Machines \\
   Massachusetts Institute of Technology \\
   Cambridge, MA 02139 \\
   \texttt{kappa666@mit.edu} \\
   \AND
   Nishka Pant \\
   Center for Brains, Minds and Machines \\
   Massachusetts Institute of Technology \\
   Cambridge, MA 02139 \\
   \texttt{npant@mit.edu} \\
   \And
   Tomaso Poggio \\
   Center for Brains, Minds and Machines \\
   Massachusetts Institute of Technology \\
   Cambridge, MA 02139 \\
   \texttt{tp@ai.mit.edu} \\
}

\begin{document}


\maketitle

\begin{abstract}
A convolutional neural network  strongly robust to adversarial perturbations at reasonable computational and performance cost has not yet been demonstrated. The primate visual ventral stream seems to be robust to small perturbations in visual stimuli but the underlying mechanisms that give rise to this robust perception are not understood. In this work, we investigate the role of two biologically plausible mechanisms in adversarial robustness. We demonstrate that the non-uniform sampling performed by the primate retina and the presence of multiple receptive fields with a range of receptive field sizes at each eccentricity  improve the robustness of neural networks to small adversarial perturbations. We verify that these two mechanisms do not suffer from gradient obfuscation and study their contribution to adversarial robustness through ablation studies.

\end{abstract}

\section{Introduction}
While modern convolutional neural networks (CNNs) have demonstrated remarkable performance in visual recognition, they still lag severely behind human vision on several tasks. Notably, despite intense effort in recent years, these models remain vulnerable to adversarial perturbations\cite{AdvPerturbations}, even on standard datasets (e.g. ImageNet). In contrast, human vision appears to be particularly robust to small perturbations in visual stimuli.

Simultaneously, an increasing body of computational and experimental work demonstrates the suitability of these artificial neural networks as models of the primate brain's ventral visual stream \cite{BrainScore, NeuralPopulationControl, GoalDrivenModelsForCortex, alex2017eigendistortions}. This suggests an opportunity to investigate if aspects of biological vision, that are not yet incorporated into current computer vision models, improve adversarial robustness. If the approach is successful, it will improve the robustness of engineered models and provide strong support for artificial neural networks being good models of primate vision. If the proposed approach does not find any way to modify the existing models to reach human-level robustness, it could demonstrate that the broad family of feedforward deep networks should be rejected as models of primate vision.



{\bf Engineering significance}
Robust real-world adversarial examples that fool models across a wide range of views, angles and lighting have been demonstrated \cite{AdvTurtle}. CNNs are increasingly being considered for use in safety critical applications (e.g. autonomous vehicles), thus the dramatic sensitivity of state-of-the-art models to tiny perturbations in otherwise benign inputs has clear security implications. Proposed methods for improving adversarial robustness have generally approached the problem from either a prevention or detection perspective. Some of the proposed mechanisms include defensive distillation \cite{DefensiveDistillation}, feature squeezing \cite{FeatureSqueezing}, defensive randomization \cite{DefensiveRandomization}, adversarial training \cite{MadryAdvTrain}, among many others. Faithfully evaluating adversarial robustness has proven to be non-trivial \cite{GradientObfuscation, EvalGuidelines, DeepMindAdvRisk}, with many proposed methods eventually being shown to be ineffective at making models robust \cite{CWPGD, GradientObfuscation, BayesianDefense, BypassFeatureSqueeze}. Adversarial training remains the most promising at increasing the robustness of models. However adversarial training comes with several downsides, ranging from significantly increased computational cost, to preferential robustness against adversarial attacks that the model was trained on \cite{BayesianDefense}, to decreased standard accuracy \cite{RobustVSAcc} and ineffectiveness at improving robustness on examples in the low density regions of the training data distribution \cite{blindspotattack}. Mechanisms that help bring the models closer to human-level robustness and alleviate this tension between performance and security would have broad implications for the application of these networks.

The learning of feature representations that are particularly well suited to visual recognition tasks is a key characteristic of CNNs that has led to their widespread use. However, the features learned by current models that perform well in standard datasets can vary significantly from features that are meaningful to humans \cite{AdvFeatures}. Recent studies have demonstrated that the quality of these learnt representations are improved in adversarially robust models. In particular, many aspects of feature visualization and manipulation were better aligned with our notion of visual perception in robust models \cite{RobustFeatureManipulate, RobustFeatureRepresentations}. This suggests that adversarial robustness holds intrinsic value, beyond security implications and potentially even at the cost of standard accuracy, as a prior for helping models learn representations that are more human meaningful and interpretable.

{\bf Biological significance}
Recent studies have argued that artificial neural networks are suitable models of biological vision due to the similarity between internal representations in CNNs and in the primate brain\cite{BrainScore, GoalDrivenModelsForCortex, NeuralPopulationControl}. In this paper, we explore two biological mechanisms that are not captured in current deep learning models of vision.  

The uneven distribution of cones in the primate retina results in non-uniform spatial sampling of visual stimuli. The density of sampling is highest at a fixation point on an image and decreases with distance from the fixation point. In contrast, standard CNNs  accept images sampled in a uniform square grid. Previous studies have demonstrated that incorporating the non-uniform sampling performed by the primate retina into standard networks improves predictivity of neural sites in the primate V4 cortical area, allows for better neural population control via controller images \cite{NeuralPopulationControl} and helps in generating adversarial examples that impact the accuracy of time-limited humans \cite{AdvHumans}. The first mechanism that we investigate incorporates information across multiple retinal fixation points.

The receptive field size along the primate visual stream increases with eccentricity \cite{Freeman2011, Gattass1981, Gattass1988}. From a sampling perspective, this translates to several scale-space image fragments centered on a fixation point in an image. Previous studies have argued for and demonstrated the computational role of pooling over the scale-space fragments in invariant visual recognition \cite{ECNN_MTHEORY}. This second ``multiple scales'' mechanism allows for translation, scale and clutter invariance to be incorporated into neural networks \cite{ECNN_SCALE, ECNN_CROWDING, yena}, with much lower sample complexity than standard data augmentation methods. We investigate incorporating information across multiple "cortical fixations".

{\bf Contributions}
We demonstrate that two mechanisms inherent to primate vision (the ``proposed mechanisms'') consistently improve the adversarial robustness of neural networks to small adversarial perturbations across a range of PGD variants, hyperparameters and adversarial criteria (by about 0\% to 30\% for $\epsilon \leq $ 0.02). One of the mechanisms (retinal fixations) improves robustness at almost no cost to standard performance (+1.62\% on ImageNet). Through ablation studies, we also identified the key features of each mechanism that contribute to robustness. Our results suggest that biologically inspired mechanisms are promising candidates for improving robustness of standard neural networks. It also probes more broadly and provides support for the hypothesis that artificial neural networks are good models of biological vision.

\section{Methods}

{\bf Datasets}
The experiments were spread across 4 datasets. CIFAR10 is a small, standard dataset \cite{CIFAR} and was used to benchmark results against other published results on adversarial robustness. The proposed mechanisms excel with images of much higher resolution. The majority of the experiments in this work were thus carried out on ImageNet10 and some experiments were repeated on ImageNet100 and ImageNet to study the scalability of the proposed mechanisms.

The ImageNet dataset \cite{IMAGENET_dataset} offers high resolution images split into 1000 classes that span a range of breadth (e.g. airliner, strawberry, etc) and depth (e.g. mud turtle, leatherback turtle, etc). 10 classes were hand picked to construct the ImageNet10 dataset. The classes were chosen to be visually distinct and of natural objects (e.g. pandas, snakes, etc.). To construct the ImageNet100 dataset, 100 classes were randomly chosen. Images in the ImageNet dataset are of varying dimensions. To standardize, all images in the ImageNet10, ImageNet100 and ImageNet datasets were formed only from the central 320x320 regions of the original images. For the full ImageNet dataset, in the interest of keeping the compute time for experiments reasonable, models were trained on the full training set but robustness evaluations were carried out on a test set downsampled to 5 images per class (totalling 5000 images for the 1000 classes).
\begin{figure*}[t!]
\begin{center}
  \includegraphics[width=\textwidth]{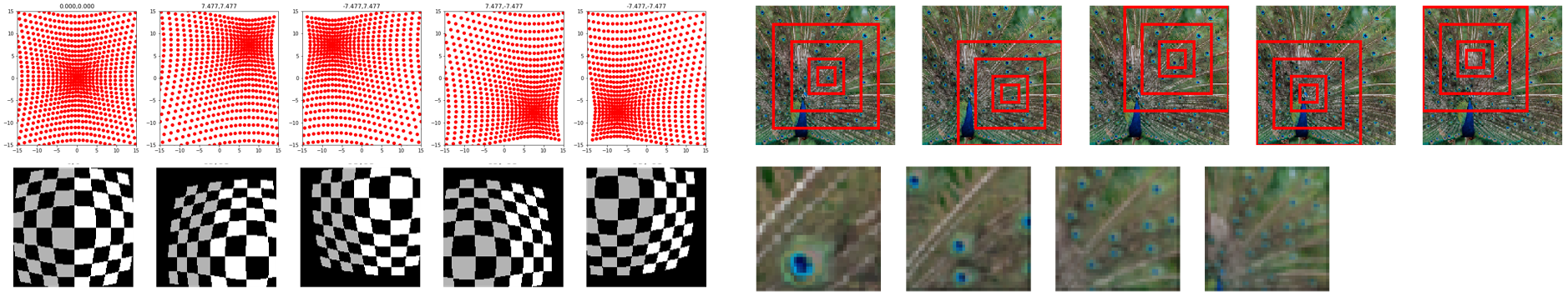}  
  \caption{Left Top: Distributions of sampling points for 5 different retinal fixations (center, top right, top left, bottom right, bottom left). Red dots represent pixels that would be sampled from the original standard image to form the retina sampled image. Left Bottom: Effect of retina sampling on an image of a flat checkerboard. Images presented were re-sampled at 5 different fixation points (same as above). Right Top: Shown in red, the centering of scale-space fragments on
5 different cortical fixations on the image (center, bottom right, top
right, bottom left, top left). Scale-space fragments for ImageNet
images were of the sizes 40x40, 80x80, 160x160, 240x240. Right Bottom: The resulting 4 scale-space fragments from 1 fixation point. For a single fixation point, the 4 scale-space fragments result from crops of varying sizes but are all Gaussian
downsampled to the size of the smallest (40x40).}
  \label{fig:sampling_summary}
\vspace{-0.79cm}
\end{center}
\end{figure*}

\begin{figure*}[t!]
\begin{center}
  \includegraphics[width=\textwidth]{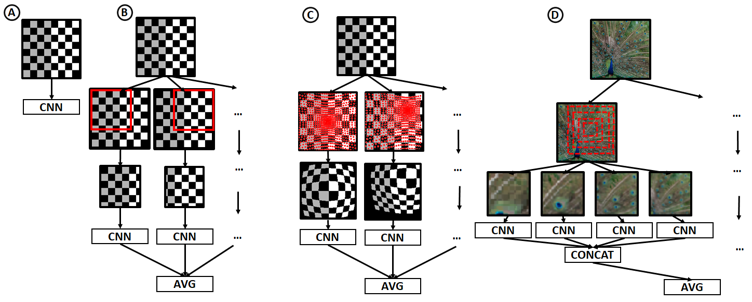}  
  \caption{Model A (baseline): standard CNN. Model B (baseline): ``coarse fixations''. Model C (effect): ``retinal fixations''. Model D (effect): ``cortical fixations''. Red marks on images in second row indicate approximate action of mechanisms centered on a fixation point. 2 fixation points visualized for Models B and C, 1 fixation point visualized for model D.}
  \label{fig:methods_summary}
\vspace{-0.6cm}
\end{center}
\end{figure*}

{\bf Biologically inspired mechanisms}
The first mechanism is the non-uniform spatial sampling of visual stimuli by the photoreceptors.  In the primate retina, the density of cones is maximum at the center of the fovea and decreases with eccentricity. The maximum density corresponds to a sampling distance of about $27$ seconds of arc between adjacent cones, which corresponds almost exactly to the Shannon sampling limit imposed by the diffraction limited optics with a cutoff around $60$ cycles/degree. The implementation of the retinal sampling was taken from \cite{NeuralPopulationControl}, where it was tuned to mimic the exponentially decreasing density of cones with eccentricity. We adapted their implementation to work with arbitrary fixation points on an image (Top Left and Bottom Left in Figure \ref{fig:sampling_summary}, Model C in Figure \ref{fig:methods_summary}). 

 The size of receptive fields also varies with eccentricity, presumably to avoid aliasing \cite{Freeman2011, Gattass1981, Gattass1988}.  The second mechanism is that  V1 neurons show a range of scales at each eccentricity \cite{ECNN_MTHEORY}. The main computational reason for this non-uniform sampling and the existence of a set of spatial scales is to enable processing of images with translation invariance -- small but growing with larger receptive fields -- and a large range of scale invariance. We assume here the estimate by \cite{Marretal1980} with $5$ ``frequency channels'' having in the fovea receptive field with a diameter of $2s=$ 1'20'', 3.1', 6.2', 11.7', 21', covering a range of roughly $1$ to $20$. Following from previous studies \cite{ECNN_MTHEORY, ECNN_SCALE, ECNN_CROWDING}, we implemented the estimates for the set of scale-space fragments in V1 (i.e. the frequency channels estimated by \cite{Marretal1980}) by taking multiple crops that are progressively larger by about a factor of 2 and Gaussian downsampling all the crops to the dimensions of the smallest crop (Top Right and Bottom Right in Figure \ref{fig:sampling_summary}, Model D in Figure \ref{fig:methods_summary}).

{\bf Models}
Models for the CIFAR10 and ImageNet datasets were based off the standard CIFAR ResNet-20 and ImageNet ResNet-18 architectures \cite{EXTREMELYDEEP_MS_2015}. 

We used a standard model and a ``coarse fixations'' model as baselines (Model A and Model B in Figure \ref{fig:methods_summary}). The coarse fixations model approximates crudely the effect of fixations by applying a standard network to different image regions. This is effectively the same as standard 10-crop testing in \cite{EXTREMELYDEEP_MS_2015} without image flipping.

The ``retinal fixations'' model applies a standard network to an image that is non-uniformly sampled as previously described (Model C in Figure \ref{fig:methods_summary}). In the ``cortical fixations'' model, the standard ResNet architecture is split over multiple branches that process the scales at each eccentricity independently. The number of filters for each branch were chosen such that the total compute cost of a forward pass of all the branches costs about the same as a standard ResNet. The striding in early layers of the ResNet architecture was adjusted to account for the small dimensions of each scale-space fragment. We used an auxiliary loss during training to encourage information across all the scales to contribute to the model prediction (Model D in Figure \ref{fig:methods_summary}).

During evaluation time, information was incorporated across fixations by averaging the model logits from 5 fixation points for each mechanism (top left, top right, bottom left, bottom right, middle).

{\bf Training}
The models for coarse, retinal and cortical fixations depend on a fixation point on the image to center the mechanism on. During training, a fixation point is randomly sampled from the valid set of fixations for the mechanism. This is in contrast to behavior at evaluation time, when information across fixations is incorporated by averaging model logits over 5 pre-determined fixation points for each mechanism. Image augmentation was standardized across all models and datasets (random crops and random left/right flips). No color augmentation was used.


Models for CIFAR10 and ImageNet10 were both trained with an ADAM optimizer with $\beta_1$ = 0.9, $\beta_2$ = 0.999 and an initial learning rate of 0.001. CIFAR10 Models were trained for 200 epochs with a batch size of 180 and a fixed learning schedule (decay from initial by 0.1, 0.01, 0.001, 0.0005 at epoch 80, 120, 160, 180). Models for ImageNet10 were trained for 400 epochs with a batch size of 64 and a fixed learning schedule (decay from initial by 0.1, 0.01, 0.001, 0.0005 at epoch 160, 240, 320, 360).  Models for ImageNet100 and Imagenet used an SGD optimizer with weight decay of 0.0001, momentum 0.9, initial learning rate of 0.1, and a batch size 256. Models for ImageNet100 were trained for 130 epochs with a fixed learning schedule (decay from initial by 0.1, 0.01, 0.001, 0.0005 at epoch 30, 70, 90, 120).  Models for ImageNet were trained for 90 epochs with a fixed learning schedule (decay from initial by 0.1, 0.01, 0.001 at epoch 30, 60, 80).

{\bf Adversarial robustness}
We followed previously proposed guidelines \cite{GradientObfuscation, EvalGuidelines, DeepMindAdvRisk} when evaluating robustness. The adversarial attacks used were as implemented in the Python package Foolbox \cite{foolbox}.

Projected Gradient Descent (PGD) \cite{PGD} was motivated by previous works as a universal first order adversary that provides a suitable security guarantee against first-order attacks \cite{MadryAdvTrain, MinimalExamples}. In this work, the adversarial robustness of the proposed mechanisms was investigated primarily with PGD. PGD ($L_{\infty}$ variant)  prescribes the generation of adversarial examples with the following iterative scheme: 
\begin{equation*}
x_{adv,i} = \textrm{CLIP}_{x,\epsilon}(x_{i-1} + \lambda \textrm{SIGN}(\nabla_x L(...))), \qquad x_{adv,0} = x_{original},
\end{equation*}
where $i$ is the count of iterations, $\textrm{CLIP}$ is an operation that clips $x$ back to the permissible set, $\lambda$ is the step size and $\nabla_x L(...)$ is the gradient of the relevant loss function for the attack. The gradient was always fully propagated through the proposed mechanisms.

When setting a maximum perturbation size $\epsilon$ with PGD, we need to choose a distance metric. Various $L_{p}$ norms of the distance from the adversarial image to the original image are usually employed,
$|x_{adv} - x_{original}|_{p}$, where typically $p=2$ or $p=\infty$ \cite{CWPGD}. The usual distance metrics are not necessarily well aligned with human perceived similarity. There is ongoing work in that area that would be interesting to consider in the future \cite{adv_perceptual}. We mostly used $L_\infty$ PGD and $L_2$ PGD in our experiments, but also checked robustness to $L_1$ PGD and the fast gradient sign method (FGSM) \cite{foolbox}.

We evaluated the robustness of the models by studying how the accuracy varied as $\epsilon$, the maximum perturbation size, was increased. We calculate the accuracy as $1$ - (num. naturally misclassified + num. adversarial examples). Whether an image is naturally misclassified was always determined by whether the true class was the most likely class predicted by the model, regardless of the adversarial criteria. Adversarial attacks were then only run against images that were not naturally misclassified. In most of the experiments, we set the step size  $\lambda$ to $\epsilon/3$ and ran 5 iterations. This allows PGD to reach the edge of the permissible set and explore the boundary while keeping compute time for the experiments reasonable. We also conducted some experiments with 20 iterations and step sizes $\lambda$ of $\epsilon/3$ and $\epsilon/12$, and tried setting $\lambda$ dynamically with an ADAM optimizer.

\section{Results}
We compared the test classification error to study standard performance. Here, we are referring to the performance of the models on the unperturbed natural test sets as the standard performance.
\begin{figure*}
  \includegraphics[width=\linewidth]{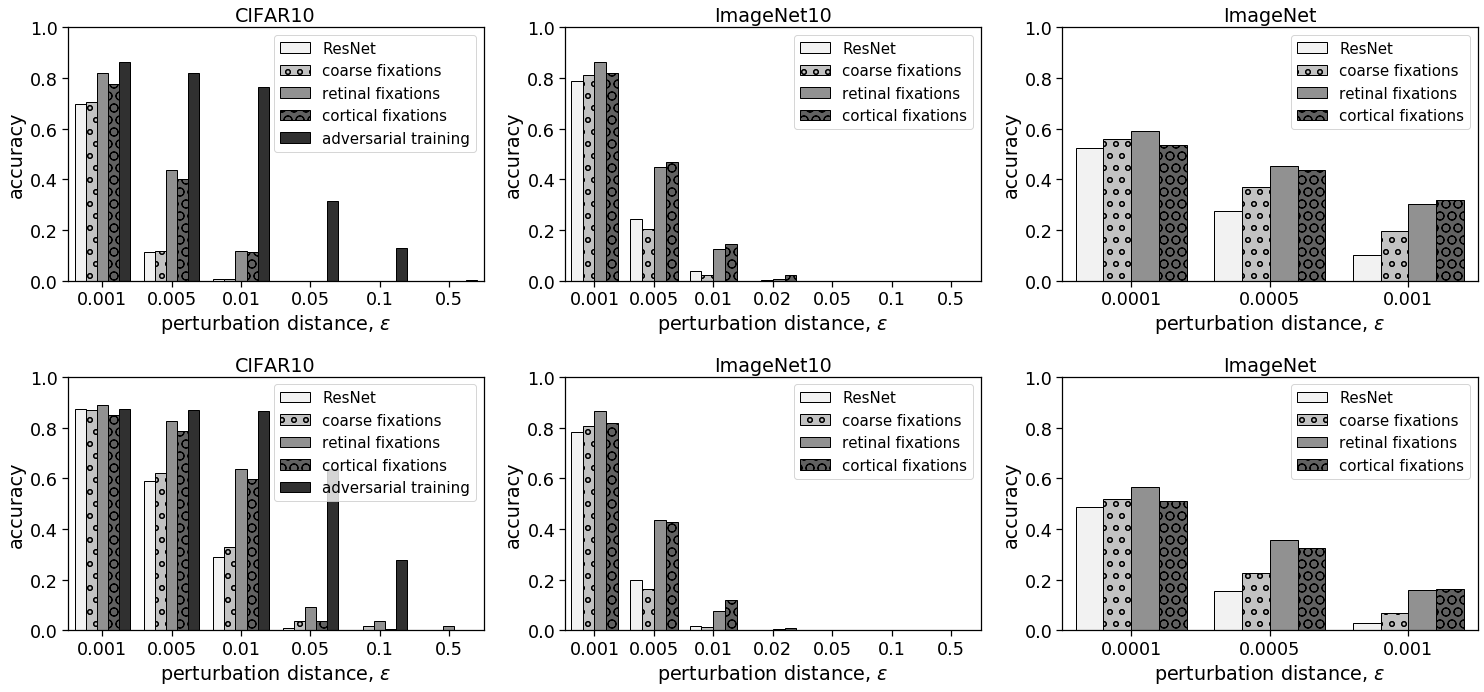}
    \caption{Robustness of the models on various datasets with increasing PGD perturbation budget. Top Row: robustness to 5-step $L_\infty$ PGD standard adversarial examples. Bottom Left: robustness to strongly misclassified adversarial examples, Bottom Middle: robustness to 20-step PGD, Bottom Right: robustness to $L_2$ PGD.}
  \label{fig:improvement_single_tests}
\vspace{-0.4cm}
\end{figure*}
The coarse fixations model generally outperformed the standard ResNet models, which is in line with previous studies \cite{EXTREMELYDEEP_MS_2015}. The coarse fixations model underperformed on CIFAR10 likely due to the small size of images.
\begin{table}[ht]
\caption{Standard TOP-1 performance of baselines and proposed models on various datasets.}
\vskip -0.1in
\label{sample-table}
\begin{center}
\begin{small}
\begin{sc}
\begin{tabular}{lccccr}
\toprule
model &  CIFAR10  &      ImageNet10 	& ImageNet100 & ImageNet \\
\midrule
   standard ResNet &    88.13\% &   89.4\% & 76.80\% & 59.46\%  \\
  coarse fixations &    87.70\% &  {\bf 91.2\% } & {\bf 78.42\% } &   61.28\% \\
   retinal fixations &    {\bf 88.88\% }&  90.2\% & 78.22\% &  {\bf 62.90\%}\\
 cortical fixations &    85.16\% &  88.6\%   & 73.62\% & 56.32\% \\
\bottomrule
\end{tabular}
\end{sc}
\end{small}
\end{center}
\vskip -0.1in
\end{table}
The retinal fixations model performed about as well as the best baseline model (worse by only 0.2\% on ImageNet100 and better by 1.62\% on ImageNet) but the cortical fixations model decreased standard performance (underperforms by 4.96\% on ImageNet). The relative performance of the models was fairly consistent across datasets.  The performance penalties for the cortical fixations model were likely still less than for adversarial training. For example, $L_{\infty}$ PGD adversarial training on CIFAR10 decreases standard performance by about 7.9\% \cite{MadryAdvTrain}.
{\bf 3.1 \ \ \ \ Adversarial robustness}

\begin{figure*}[h!]
\includegraphics[width=\linewidth]{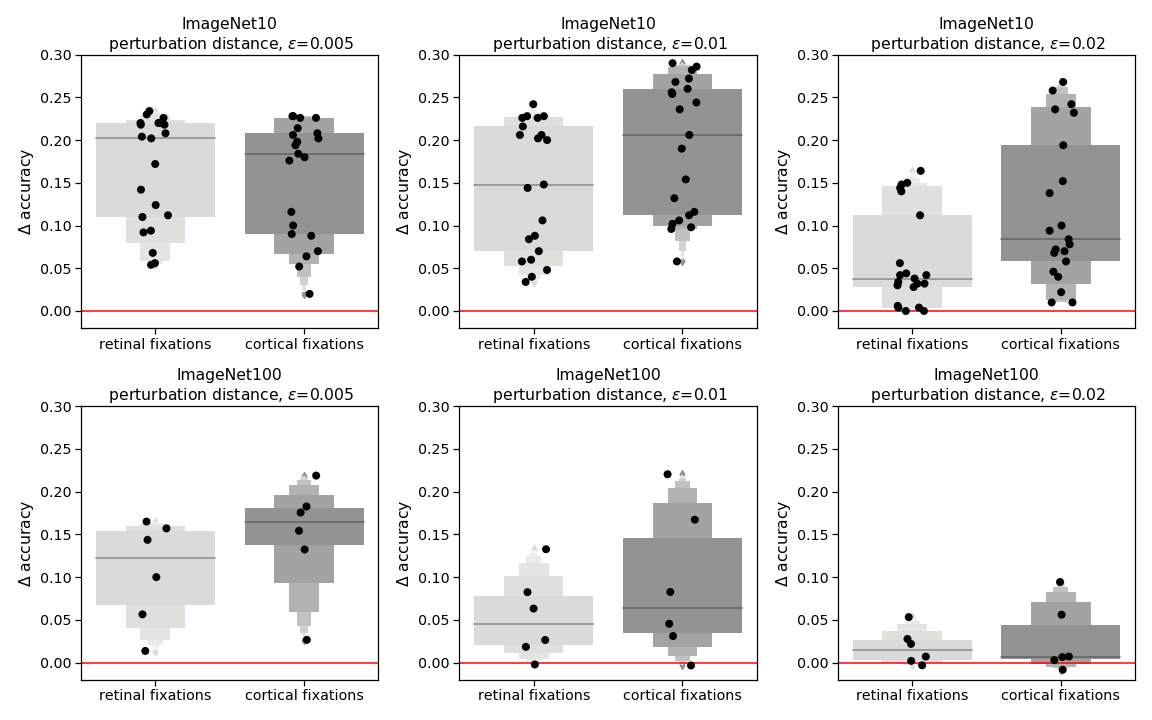}
    \caption{Improvement in accuracy of proposed models from the best baseline at small $\epsilon$. Each dot represents a different attack configuration. Top row: ImageNet10. Bottom row:  ImageNet100.}\label{fig:improvmenet_aggregate}
\vspace{-0.20cm}
\end{figure*}

The adversarial robustness of the proposed mechanisms was evaluated by studying the change in accuracy of the models with increasing perturbation budget. The experiments span a range of PGD variants, hyperparameters and adversarial criteria (Figures \ref{fig:improvement_single_tests}, \ref{fig:improvmenet_aggregate}).

The biologically inspired mechanisms consistently improved adversarial robustness to small perturbations across almost all experiments on all datasets (see Figure \ref{fig:improvmenet_aggregate}). The extent of improvement varied depending on the attack variants, hyperparameters and the dataset used. The retinal fixations and cortical fixations models improved robustness to about the same extent, with the best model varying across experiments. At larger perturbations, the mechanisms did not significantly impact robustness.

For small perturbations on ImageNet10, we observed a greater improvement in accuracy for the proposed models from the best baseline when using PGD as opposed to FGSM. This suggests that the mechanisms are effective at improving robustness, with the retinal and cortical fixations  models performing proportionately better against stronger attacks that more thoroughly probe the robustness. The improvement in accuracy was relatively unchanged when increasing the number of iterations of PGD from 5 to 20 with step size $\lambda$ of 0.1 or 0.025. In general, the proposed models showed greater improvement in accuracy under $L_\infty$ PGD versus $L_2$ PGD, with the least improvement with $L_1$ PGD. Further studies could be conducted on variations in improvement with attack variants and hyperparameters to better understand the contribution of the mechanisms to adversarial robustness.
\vspace{-0.15cm}


{\bf Confident misclassifications}
On ImageNet10, the robustness to standard misclassifications (true class not the most likely predicted class) was compared to the robustness to confident misclassifications (true class not in top 3 for untargeted attacks and adversarial class predicted with probability $>$80\% for targeted attacks). Generally,  the proposed mechanisms were more robust to confident mistakes than standard mistakes for untargeted attacks. The robustness to confident and standard mistakes for targeted attacks was about the same.

In a targeted setting, the loss used for PGD pushes towards a confident misclassification (objective is to increase probability of adversarial class) whereas in an untargeted setting, the loss does not explicitly push towards confident misclassifications (objective is to minimize true class, this can be accomplished either by increase probability of only 1 or any number of other classes). This suggests that the proposed models improve robustness to confident misclassifications as confident mistakes were less likely to occur unless they were explicitly optimized for.

{\bf Adversarial Training}
On CIFAR10, the robustness of the models was benchmarked against an adversarially trained model available online from a previous study \cite{MadryAdvTrain}. The model uses a variant of the standard ResNet architecture, w32-10 ResNet. We benchmarked the robustness of the models with $L_\infty$ PGD, which is also the attack that was used for adversarial training (Figure \ref{fig:improvement_single_tests}).

The robustness of the retinal fixations and cortical fixations models was consistently better than the baselines but worse than adversarial training. The accuracy of the models that were not adversarially trained dips to approximately 0\% for perturbations of size $\epsilon$ = 0.05 and higher. The proposed models made far fewer confident misclassifications  than the baselines, with the robustness of the proposed models to confident misclassification at $\epsilon$ = 0.005 approaching adversarial training.

%

%

{\bf 3.2 \ \ \ \ Ablation experiments}

\begin{figure*}[h!]
\begin{center}
\vspace{-0.3cm}
  \includegraphics[trim =  15 15 10 15, width=\linewidth, clip]{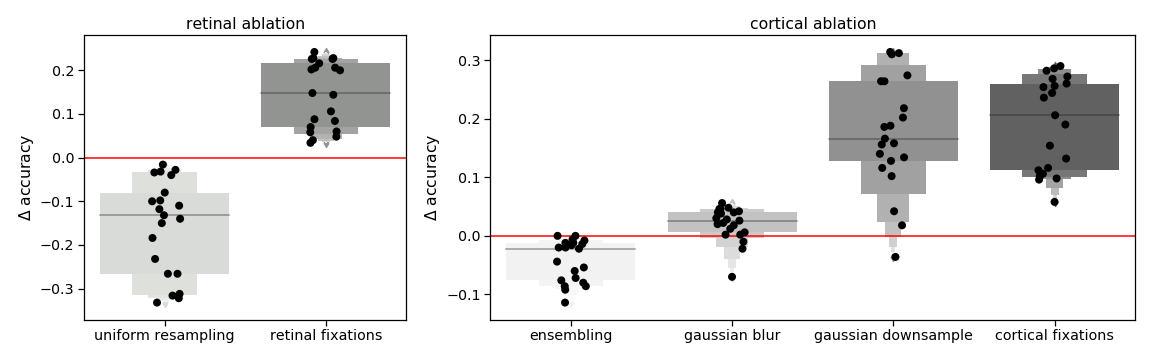}
      \caption{Improvement in accuracy of ablated models over the best baseline model under $L_\infty$ PGD with $\epsilon$ = 0.01. Each dot represents a different attack configuration/criteria.}
  \label{fig:robustness_ablation}
\vspace{-0.5cm}
 \end{center}
\end{figure*}

We performed experiments with a series of ablated models to study the contribution of various aspects of the proposed mechanisms to robustness. The experiments were conducted on ImageNet10 and span a range of PGD variants, hyperparameters and adversarial criteria (Figure \ref{fig:robustness_ablation}).  All ablated models employed a mechanism centered on fixations. As before, we incorporated information across fixations at evaluation time by averaging the model logits from 5 pre-determined fixation points in the image.

{\bf Retinal ablation}
Non-uniform retinal sampling effectively sub-samples and up-samples the image. The pixels that get sub-sampled and up-sampled are chosen based on their distance to a point of fixation on the image. There are several factors that could potentially account for the improved robustness of the retinal fixations model, such as the sub-sampling or the sub-sampling combined with up-sampling or the non-uniformity of the sampling.

The retinal fixations model consistently outperformed the coarse fixations model in robustness tests. This indicates that simply sub-sampling, or applying a network on different image regions, does not improve robustness. We studied an ablated model that performs uniform sub-sampling (exactly as the coarse fixations model) followed by up-sampling. This ablated model (``uniform resampling'' in Figure \ref{fig:robustness_ablation}) consistently underperformed compared to the best baseline model. This indicates that uniformly sub-sampling and up-sampling an image does not improve (actually worsens) robustness.

This leaves the non-uniformity of the sampling as the main contributing factor to improving robustness. We speculate that the non-uniform sampling appropriately conditions the model to learn larger scale features during the training process and that these features are more robust to adversarial perturbations. This could be investigated further by studying the extent of robust versus non-robust features learnt by the model as in \cite{AdvFeatures}.

{\bf Cortical ablation}
The cortical fixations model breaks an image into multiple scale-space fragments centered on a given fixation point. These fragments are constructed with a uniform square crop of increasing size followed by a Gaussian downsample of all fragments to the dimensions of the smallest fragment. The fragments are then fed through separate branches of a network and the linear classifier at the end of the network uses information from all branches. Potential factors contributing to the improved robustness include Gaussian blurs on sub-samples of the image, Gaussian downsampling on sub-samples of the image, classifying with a fixed computational budget distributed over separate branches of a network that combine their output, or a combination of the factors.

Classifying with a branched network (``ensembling'' ablated model in Figure \ref{fig:robustness_ablation}) worsened or at best did not change the robustness of the models. The ``Gaussian blur'' ablated model performs the sub-sampling exactly as the coarse fixations model, but applies a Gaussian blur on the images. This ablated model was more robust than the baselines but only to a small extent. It has been noticed previously that denoising operations (e.g. Gaussian blur) contribute to adversarial robustness \cite{AdvTrainDenoiseFB, AvgFilter}.

The ``Gaussian downsample'' ablated model is effectively the branch of the cortical fixations model that corresponds to the largest receptive field size. It performs sub-sampling similar to the coarse fixations model and applies a Gaussian downsample. This ablated model was more robust than the baselines but less robust than the cortical fixations model. This suggests that increasing reliance on large scale features in an image for classification partially contributed to the improved robustness of the cortical fixations models. 

This leaves the ensembling across different scale-space fragments as the remaining contributing factor to the robustness of the cortical fixations model. It is possible that the fragments are sufficiently de-correlated to make ensembling effective, unlike in the case of the ``ensembling'' ablated model. We tried several versions of the cortical fixations model that incorporated information across the different scale-space fragments in various ways. Some  methods (e.g. pooling across branches) outperformed the current cortical fixations model in several experiments. This suggests that issues remain with properly incorporating features across the scale-space fragments in the cortical fixations model.

{\bf Gradient obfuscation}
A common pitfall when evaluating adversarial robustness is gradient obfuscation \cite{GradientObfuscation}. By masking the gradient, models can appear to be robust under white-box attacks (e.g. PGD) which use the gradient to construct an adversarial example. We verified that the proposed mechanisms did not suffer from gradient obfuscation through previously suggested experiments \cite{EvalGuidelines}.

{\bf 0\% Accuracy on unbounded attacks}
Unbounded iterative PGD should always be able to bring the accuracy of models down to 0. We achieved 0\% accuracy in out models at $\epsilon=0.5$ under $L_{\infty}$ PGD and $L_2$ PGD on ImageNet100.

{\bf White-box attacks outperform black-box attacks}
The gradient propagated through the models should be more useful for constructing adversarial examples than a coarse approximation of the gradient. Adversarial examples were constructed by estimating the gradient of the retinal fixations model and cortical fixations model with a standard ResNet model (transfer attack). The examples generated with the transfer attack (black-box attack) heavily underperformed compared to PGD (white-box attack). In further experiments, PGD (white-box attack) also outperformed boundary-attack \cite{BoundaryAttack} and backward pass differentiable approximation \cite{GradientObfuscation}.

{\bf Iterative attacks outperform single-step attacks}
Following the direction of gradient descent should generally be useful for generating stronger adversarial examples. The accuracy of our models was strictly greater or equal under FGSM, effectively a one-step PGD attack, than under PGD.

{\bf Expectation over random transformations}
The coarse, retinal and cortical fixations models can be used as stochastic models. The transformation performed by the mechanisms at training time was random. However, at evaluation time, the transformations were pre-determined and fixed. Thus, there was no need for taking expectations over the transformation in almost all of our experiments. We conducted a single set of experiments with random transformations at evaluation time. For that experiment, we took an expectation of the gradient from 5 random samples of the transformation.

\vspace{-0.2cm}
\section{Conclusion}
In this work, we showed that two key features of primate vision -- foveation due to non-uniform distribution of cones in the retina and multiscale filtering because of receptive fields of different sizes in V1 at each eccentricity -- consistently improve the robustness of neural networks to small adversarial perturbations. These mechanisms have negligible computational effect and one of them (non-uniform sampling) improves robustness at almost no cost in recognition performance.

However, these mechanisms did not improve robustness to large perturbations. Preliminary inspection of the adversarial examples leads us to speculate that the inability of human perception to notice that the adversarial images are different from the normal ones breaks roughly around or above the transition in our experiments from ``small'' to ``large'' perturbation. In that case, our results suggest that the two mechanisms we identified may partially explain the robustness of primate vision to ``small'' perturbations, while an additional, separate mechanism -- akin to an anomaly detector -- may detect the presence of  ``large'' perturbations. 

In summary, our results lend  support to the hypothesis that artificial neural networks are suitable as a core model of object recognition in primates. They also suggest that implementing further biologically inspired mechanisms is a promising direction for increasing adversarial robustness  without sacrificing standard performance.

	\small
	\subsubsection*{Acknowledgments}
	We thank Jim DiCarlo for illuminating discussions. Part of the funding is from
	Center for Brains, Minds and Machines (CBMM), funded by NSF STC award
	CCF-1231216, and part from Lockheed Martin.
	
	\normalsize

\section*{Broader impact}
In terms of ethical aspects and future societal consequences, achieving adversarially robust models
is of critical importance to deployment of autonomous technologies we can trust. Among some of the positive outcomes 
we could count autonomous vehicles immune to misleading by malicious agents, better trust in medical imaging applications
and any other discipline to which AI can be applied and which requires safety guarantees. This naturally also comes with negative impacts,
from reducing jobs available to humans, to
potentially making surveillance technologies much more difficult to avoid, allowing authoritarian regimes a much tighter control
over their citizens, as well as enabling progress in autonomous weapon systems. Additionally, every defense against adversarial attacks
that has been so far proposed has eventually been found to be vulnerable. If such vulnerabilities are found in increasingly more
accurate models of primate vision, this could suggest the possibility of existence of dynamically changing adversarial attacks that would fool humans,
leading to potentially new camouflage technologies.

\small
	\newpage
	\bibliographystyle{unsrt}
	\bibliography{Boolean}
	\normalsize
	\newpage
	\begin{center}
		{\bf \large Appendix}
	\end{center}

\section{Datasets}

\subsection{Preprocessing}

All images $x$ were centered on the $0$ to $1$ scale by $x/255$. 

Images in ImageNet10 and ImageNet100 were formed from the central 320x320 regions of images from ImageNet. If the image dimensions for either height or width (or both) were less than 320, we upsampled the image with a bilinear filter with a constant aspect ratio so that both dimensions were at least 320x320.

We conducted a set of preliminary experiments on ImageNet10 with training images (not test images) trimmed to bounding boxes. Results from these preliminary experiments were not reported in the paper but we report the results here in the supplementary materials. The standard bounding boxes were used as provided with the ImageNet dataset. If images had 0 bounding boxes, they were discarded for this dataset. If images had 1 bounding box, they were trimmed to the bounding box before the standard central 320x320 crop described above was performed. If images had more than 1 bounding box, we only kept the first bounding box.

\subsection{ImageNet10}
Classes comprising the dataset are listed below:
\begin{enumerate}
\itemsep0em 
	\item Snake: n01742172 boa constrictor
	\item Dog: n02099712, Labrador retriever 
	\item Cat: n02123045, tabby
	\item Frog: n01644373, tree frog
	\item Turtle: n01665541, leatherback turtle
	\item Bird: n01855672 goose
	\item Bear: n02510455 giant panda
	\item Fish: n01484850 great white shark
	\item Crab: n01981276 king crab
	\item Insect: n02206856 bee
\end{enumerate}

\subsection{ImageNet100}
Classes comprising the dataset are listed below:

['n01644900', 'n02096051', 'n04366367', 'n03544143', 'n02105412',
       'n01914609', 'n02105162', 'n02132136', 'n03026506', 'n03063599',
       'n02815834', 'n07802026', 'n01968897', 'n03788365', 'n04443257',
       'n12998815', 'n02454379', 'n03991062', 'n04332243', 'n04254680',
       'n02097298', 'n07590611', 'n03680355', 'n02165105', 'n01491361',
       'n04120489', 'n03742115', 'n07880968', 'n02808304', 'n03888257',
       'n03095699', 'n01494475', 'n03673027', 'n02488702', 'n01871265',
       'n02104365', 'n02281787', 'n04118538', 'n01828970', 'n02837789',
       'n03127747', 'n04005630', 'n02115913', 'n01514859', 'n03452741',
       'n02107908', 'n01847000', 'n04200800', 'n04153751', 'n04389033',
       'n02487347', 'n02769748', 'n01843383', 'n02219486', 'n02009912',
       'n03676483', 'n02797295', 'n04417672', 'n04591157', 'n04229816',
       'n02058221', 'n03814906', 'n02097130', 'n02939185', 'n03710637',
       'n02116738', 'n04418357', 'n03775071', 'n04328186', 'n02090721',
       'n02667093', 'n03929855', 'n02089078', 'n02389026', 'n03388183',
       'n07613480', 'n02749479', 'n02174001', 'n07932039', 'n02112018',
       'n02398521', 'n04069434', 'n03838899', 'n02233338', 'n03207743',
       'n02791270', 'n02114855', 'n04204238', 'n02342885', 'n02110063',
       'n01518878', 'n02099712', 'n01704323', 'n02168699', 'n04238763',
       'n03494278', 'n03980874', 'n02097209', 'n01616318', 'n03131574']

\section{Biologically Inspired Mechanisms}

\subsection{Retinal Sampling}

Implementation for retinal sampling was adapted from \cite{NeuralPopulationControl}. Please consult their paper for full details on the sampling procedure and chosen parameters. In their work, they defined a function $g$, that mapped points from the re-sampled image ($r'$) to the original image ($r$). Their code was open-sourced at https://github.com/dicarlolab/retinawarp. As part of the transform, the pixel coordinates in the image grid ($x,y$) were mapped to polar ($r,\theta$). To work with arbitrary fixation points, we simply re-centered this transform at a fixation point instead of at the Cartesian origin. We present here a visualization of the distribution of the sampling points (Figure \ref{fig:sampling_summary}: Top Left) and some examples of retina sampled images at 5 different fixation points (Figure \ref{fig:sampling_summary}: Bottom Left).



\subsection{Cortical Sampling}

\begin{figure}[h!]\centering
\includegraphics[width=\columnwidth]{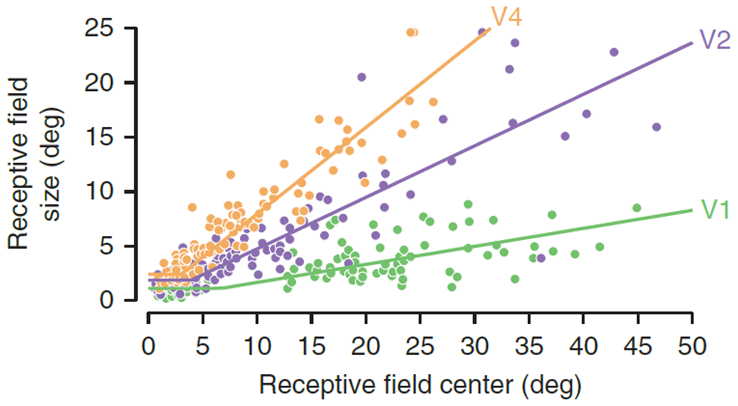}
\caption{Biological measurements in V1, V2 and V4 support a model in which the maximum scale depends on eccentricity. Adapted from \cite{Freeman2011} (original monkey data from \cite{Gattass1981,Gattass1988}).}
\label{fig:freeman}
\end{figure}

Biological measurements \cite{Gattass1981, Gattass1988} have demonstrated that in primates, the receptive field size varies with eccentricity. With the slope of the relationship between receptive field size and eccentricity also becoming steeper further down the ventral stream (Figure \ref{fig:freeman}). Following from previous studies \cite{ECNN_MTHEORY, ECNN_SCALE, ECNN_CROWDING}, we approximate this by centering multiple scale-space fragments on a point of fixation on the image, with the larger scales corresponding to larger receptive field sizes. We employed 4 scales of dimensions 40x40, 80x80, 160x160 and 240x240 that were all gaussian downsampled to 40x40 for ImageNet. For CIFAR10, we only employed 2 scales (15x15 and 30x30 that were both gaussian downsampled to 15x15) because of the significantly lower resolution of images in the dataset (32x32 in CIFAR10 vs 320x320 for ImageNet). We present here a visualization of the centering of the scale-space fragments at various fixation points in an image (Figure \ref{fig:sampling_summary}: Top Right) and some of the resulting scales (Figure \ref{fig:sampling_summary}: Bottom Right).

When making a prediction for a given fixation point, models incorporating cortical sampling have to incorporate information across the different scales. Each scale is fed to a separate CNN. These separate CNNs do not end with a dense layer. We concatenate the final latent vector from each branch and this concatenated representation was then passed through a single dense layer for classification. During training time, we used an auxiliary loss to ensure that each branch was also predictive of the label on its own. We also provide results from some preliminary experiments with alternate concatenation mechanisms (e.g. max, average pooling across scales, etc.) in the supplementary results section below.



\section{Models}

As described in the paper, we employed two baseline models ('ResNet' and 'coarse fixations') and two effect models ('retinal fixations' and 'cortical fixations'). At evaluation time, the 'coarse fixations', 'retinal fixations' and 'cortical fixations' models each center their mechanisms on 5 pre-defined fixation points (top left, top right, bottom left, bottom right, center) and combine information across fixations by averaging models logits across the fixation points.

The 'ResNet' baseline model directly feeds the full image through a standard ResNet architecture (32x32 for CIFAR10 or 320x320 for ImageNet). The 'coarse fixations' model applies a standard ResNet architecture to 5 different regions of the image (5 224x224 regions from the 320x320 image for ImageNet, centered on (0,0),(48,48),(48,-48),(-48,48),(-48,-48) or 5x24x24 from 32x32 for CIFAR10, centered on (0,0),(4,4),(4,-4), (-4,4), (-4,-4)). This acts as a very coarse approximation of foveation. At training time, the model randomly trains on a single region from all valid regions, not just the 5 to be used at evaluation time (i.e. centered on (rand(-48 to 48),rand(-48 to 48)) for ImageNet and centered on (rand(-4 to 4),rand(-4 to 4)) for CIFAR10).

The 'retinal fixations' model applies the retinal sampling at 5 different fixation points (re-samples the image, keeping dimensions the same). At evaluation time, the mechanism is centered on the points (0,0),(80,80),(80,-80),(-80,80),(-80,-80) for ImageNet and on the points (0,0),(8,8),(8,-8),(-8,8),(-8,-8) for CIFAR. At training time, the model randomly trains on a single fixation from all possible fixations (centered on (rand(-80 to 80),rand(-80 to 80) for ImageNet and centered on (rand(-8 to 8),rand(-8 to 8)) for CIFAR10).

The 'cortical fixations' model applies the cortical sampling at 5 different fixation points (re-samples the ImageNet images from 320x320 to 5x40x40 and CIFAR10 images from 32x32 to 2x15x15). At evaluation time, the mechanism is centered on the points (0,0),(40,40),(40,-40),(-40,40),(-40,-40) for ImageNet and (0,0),(1,1),(1,-1),(-1,1),(-1,-1) for CIFAR10. At training time, the model randomly trains on a single fixation from all possible fixations (centered on (rand(-40 to 40),rand(-40 to 40)) for ImageNet and centered on (rand(-1 to 1),rand(-1 to 1)) for CIFAR10).

\section{Adversarial Robustness}

\subsection{Determinism In TensorFlow}

We performed all experiments, training, etc. on TensorFlow 2.0.0. On this version, TensorFlow implements a reduced form of GPU-deterministic op functionality. We fix this to ensure deterministic operations with a patch from NVIDIA (see https://github.com/NVIDIA/tensorflow-determinism). This patch was applied only when evaluating adversarial robustness. We verified that the results from 3 experiments (coarse fixations, retinal fixations, cortical fixations models for 5 iterations of $L_{\infty}$ PGD with a targeted loss (y\_adv = y\_true + 1) with $\epsilon$=0.01 and step size of $\epsilon$ / 3) were consistent with and without this patch. See ``check determinism'' notebook in the checks folder in source code for additional details.

\subsection{Accuracy Calculation}
To calculate the accuracy under an adversarial attack, we employed the following procedure. We first evaluated the model on the natural images in the test splits for the datasets. This gives us a number of natural images that are correctly classified and a number that are misclassified (always determined by the true-class not being the top-1 class predicted by model, num\_misclassified). We then ran the adversarial attack on the correctly classified natural images, generating some number of adversarial examples that meet the desired adversarial criteria (num\_adversarial). We then calculated the accuracy under attack as 1-((num\_misclassified+num\_adversarial)/(total\_num\_images)).

\section{Results}

\subsection{CIFAR10 Benchmark}

We benchmarked our evaluation pipeline against the pipeline used in \cite{MadryAdvTrain} (see Table \ref{cifar10_benchmark}). We perform these benchmarks on the CIFAR10 test set of 10000 images. Across our experiments, our pipelines were exactly consistent (except in the one case of 20-step PGD where our results differed by 1 image on the full test set). The benchmark pipeline from \cite{MadryAdvTrain} was open-sourced at https://github.com/MadryLab/cifar10\_challenge.

\begin{table}[t]
\caption{Comparison of accuracy under PGD attacks between our pipeline and the pipeline used in \cite{MadryAdvTrain}. Top Table: 5-step $L_{\infty}$ PGD, step size of $\epsilon$/3. Bottom Table: 20-step $L_{\infty}$ PGD step size, step size of 2/255.}
\label{cifar10_benchmark}
\vskip 0.15in
\begin{center}
\begin{small}
\begin{sc}
\begin{tabular}{lcccr}
\toprule
             $\epsilon$ &    benchmark accuracy &  our accuracy \\
\midrule
 0.001 &  86.15\% &  86.149\% \\
 0.005 & 82.06\% & 82.06\% \\
 0.01 & 76.60\% & 76.60\% \\
 0.05 & 31.26\% & 31.26\% \\
 0.1 & 12.95\% & 12.949\% \\
 0.5 & 0.12\% & 0.1199\% \\
\bottomrule
\toprule
             $\epsilon$ &    benchmark accuracy &  our accuracy \\
\midrule
 8/255 &  45.72\% &  45.73\% \\
\bottomrule
\end{tabular}
\end{sc}
\end{small}
\end{center}
\vskip -0.1in
\end{table}

\subsection{All Experiment Hyperparameters}

We provide an almost complete list of hyperparameters used for the experiments presented in the paper (Table \ref{hyperparams}). Please refer to the full evaluation data provided with the source code for an exhaustive list of all relevant hyperparameters. For ImageNet10 for example, not presented in the table,  we also ran 1 experiment with 5 randomly chosen fixation points and 1 experiment where the images were initialized with random noise in an $\epsilon_{init}$ ball. Both these experiments were run with $L_{\infty}$ PGD for 5 steps with ATTACK\_STEP\_SIZE = 0.1 and CRITERION\_TAG = MISCLASSIFY\_3.

For each set of hyperparameters, the value for maximum perturbation size $\epsilon$ was scanned across  [0.001, 0.005, 0.01 , 0.05 , 0.1  , 0.5  ] for CIFAR10 and [0.001, 0.005, 0.01 , 0.02, 0.05 , 0.1  , 0.5  ] for ImageNet. 

\begin{table*}[t]
\caption{Attack hyperparameters used for evaluating the robustness of the models in most of the experiments presented in the paper. Please see full evaluation data provided with the source code for an exhaustive list. ATTACK\_ALGO refers to the algorithm used for generating the adversarial example, ATTACK\_DISTANCE\_METRIC refers to the form of $L_p$ norm used to measure distances and the $L_p$ variant of the attack algorithm, ATTACK\_ITERATIONS refers to the number of steps of the iterative algorithms, ATTACK\_STEP\_SIZE refers to a constant used internally to calculate the actual step size (actual step size: ($\epsilon$ / 0.3) * ATTACK\_STEP\_SIZE), CRITERION\_TAG refers to the loss and criteria used for the attack (TARGETED\_X: targeted attack (adv\_class = true\_class + 1, classified as adv\_class with probability at least X if X != 50, classified as adv\_class as top1 predicted class if X = 50), MISCLASSIFY\_X: untargeted attack (true class not in top-X classes predicted by model)). Top Table: For experiments on CIFAR10, Middle Table: IMAGENET10, Bottom Table: IMAGENET100.}
\label{hyperparams}
\vskip 0.15in
\begin{center}
\begin{small}
\begin{sc}
\begin{tabular}{llrrl}
\toprule
attack\_algo & attack\_distance\_metric &  attack\_iterations &  attack\_step\_size &  criterion\_tag \\
\midrule
        PGD &                     LINF &                  5 &             0.100 &    misclassify\_1 \\
                PGD &                     LINF &                  5 &             0.100 &    misclassify\_3 \\
\bottomrule
\toprule
attack\_algo & attack\_distance\_metric &  attack\_iterations &  attack\_step\_size &  criterion\_tag \\
\midrule
        PGD &                     L2 &                  5 &             0.100 &    targeted\_50 \\
        PGD &                   LINF &                 20 &             0.025 &    targeted\_50 \\
        PGD &                   LINF &                 20 &             0.100 &    targeted\_50 \\
        PGD &                   LINF &                  5 &             0.100 &    targeted\_50 \\
        PGD &                     L2 &                  5 &             0.100 &  misclassify\_1 \\
        PGD &                   LINF &                 20 &             0.025 &  misclassify\_1 \\
        PGD &                   LINF &                 20 &             0.100 &  misclassify\_1 \\
        PGD &                   LINF &                  5 &             0.100 &  misclassify\_1 \\
       FGSM &                   LINF &                  1 &            -1.000 &  misclassify\_3 \\
        PGD &                     L1 &                  5 &             0.100 &  misclassify\_3 \\
        PGD &                     L2 &                  5 &             0.100 &    targeted\_80 \\
        PGD &                     L2 &                  5 &             0.100 &  misclassify\_3 \\
        PGD &                   LINF &                 20 &             0.025 &    targeted\_80 \\
        PGD &                   LINF &                 20 &             0.025 &  misclassify\_3 \\
        PGD &                   LINF &                 20 &             0.100 &    targeted\_80 \\
        PGD &                   LINF &                 20 &             0.100 &  misclassify\_3 \\
        PGD &                   LINF &                  5 &             0.100 &    targeted\_80 \\
        PGD &                   LINF &                  5 &             0.100 &  misclassify\_3 \\
   PGD\_ADAM &                   LINF &                  5 &             0.100 &  misclassify\_3 \\
\bottomrule
\toprule
attack\_algo & attack\_distance\_metric &  attack\_iterations &  attack\_step\_size &  criterion\_tag \\
\midrule
        PGD &                     L2 &                  5 &             0.100 &    misclassify\_1 \\
        PGD &                     LINF &                  5 &             0.100 &    misclassify\_1 \\
        PGD &                     L2 &                  5 &             0.100 &    misclassify\_3 \\
        PGD &                     LINF &                  5 &             0.100 &    misclassify\_3 \\
        PGD &                     L2 &                  5 &             0.100 &    misclassify\_10 \\
        PGD &                     LINF &                  5 &             0.100 &    misclassify\_10 \\
\bottomrule
\end{tabular}

\end{sc}
\end{small}
\end{center}
\vskip -0.1in
\end{table*}

\subsection{Robustness To FGSM, PGD, PGD ADAM}

\begin{figure}\centering
\includegraphics[width=\columnwidth]{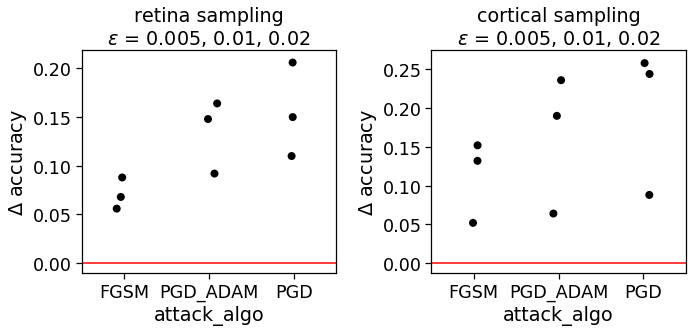}
\caption{Improvements in robustness for retinal fixations and cortical fixations models from the best baseline under various adversarial attacks at small perturbations on ImageNet10.}
\label{fig:attack_algos}
\end{figure}

We compared the robustness for the FGSM, PGD ADAM, PGD adversarial attacks using the $L_{\infty}$ distance metric, 5 attack iterations (1 for FGSM), step size of $\epsilon$ / 3 (5$\epsilon$ / 3 for FGSM), adversarial criteria of true class not in top-3 predicted classes by model at $\epsilon$ = [0.005, 0.01, 0.02]. Retinal fixations and cortical fixations models showed biggest improvements in robustness from the best baseline for PGD, and the least with FGSM (see Figure \ref{fig:attack_algos}). 

\subsection{Robustness To PGD 5 Steps, PGD 20 Steps}

\begin{figure}\centering
\includegraphics[width=\columnwidth]{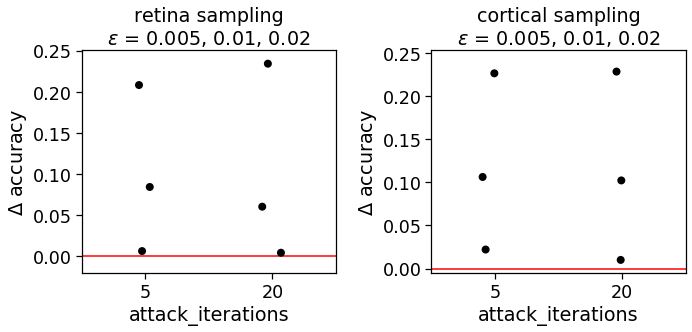}
\caption{Improvements in robustness for retinal fixations and cortical fixations models from the best baseline at varying PGD attack iterations at small perturbations on ImageNet10.}
\label{fig:attack_iterations}
\end{figure}

We compared the robustness for $L_{\infty}$ PGD at 5 and 20 iterations. We ran the attacks with a step size of $\epsilon$ /3, with the adversarial criteria of true class not top-1 predicted class by model at $\epsilon$ = [0.005, 0.01, 0.02]. Improvement in robustness of retinal fixations and cortical fixations models from the best baseline was approximately the same for 5 and 20 iterations (see Figure \ref{fig:attack_iterations}).

\subsection{Robustness To PGD Step Size $\epsilon$ /3, $\epsilon$ /12}

\begin{figure}\centering
\includegraphics[width=\columnwidth]{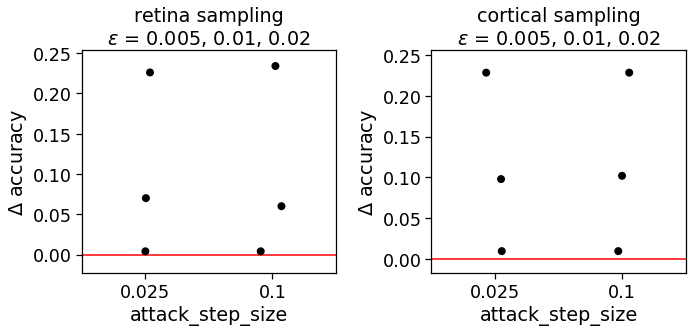}
\caption{Improvements in robustness for retinal fixations and cortical fixations models from the best baseline at varying PGD attack steps at small perturbations on ImageNet10. ATTACK\_STEP\_SIZE = 0.1, 0.025 corresponds to a PGD step size of $\epsilon/3$ and $\epsilon$ / 12.}
\label{fig:attack_stepsizes}
\end{figure}

We compared the robustness for $L_{\infty}$ PGD at step sizes of $\epsilon$ /3 (ATTACK\_STEP\_SIZE = 0.1) and $\epsilon$ /12 (ATTACK\_STEP\_SIZE = 0.025). We ran the attacks for 20 iterations, with the adversarial criteria of true class not top-1 predicted class by model at $\epsilon$ = [0.005, 0.01, 0.02]. Improvement in robustness of retinal and cortical fixations models from the best baseline was approximately the same for the 2 step sizes (see Figure \ref{fig:attack_stepsizes}).

\subsection{Robustness Under $L_{\infty}$, $L_2$, $L_1$ Metrics}

\begin{figure}\centering
\includegraphics[width=\columnwidth]{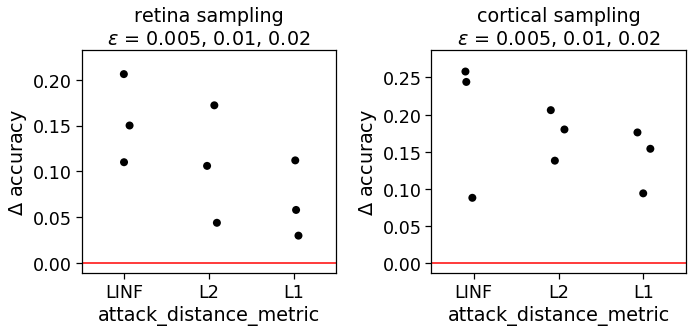}
\caption{Improvements in robustness for retinal and cortical fixations models from the best baseline using various distance metrics at small perturbations on ImageNet10.}
\label{fig:attack_distmetrics}
\end{figure}

We compared the robustness for $L_{\infty}$, $L_2$ and $L_1$ PGD run for 5 iterations with a step size of $\epsilon$ /3 and the adversarial criteria of true class not in the top-3 predicted classes by model at $\epsilon$ = [0.005, 0.01, 0.02]. Improvement in robustness of retinal and cortical fixations models was greater for $L_{\infty}$ and least for $L_1$ (see Figure \ref{fig:attack_distmetrics}).

\subsection{Robustness To Confident Mistakes}

\begin{figure}\centering
\includegraphics[width=\columnwidth]{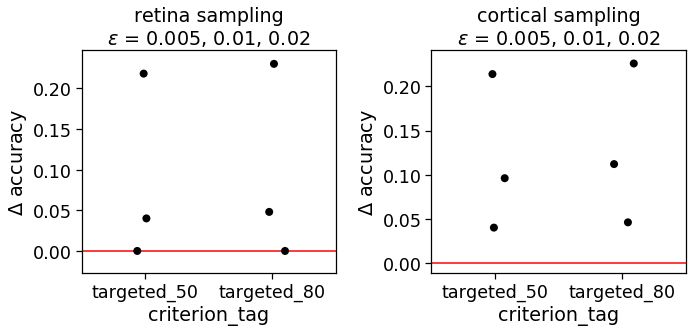}
\includegraphics[width=\columnwidth]{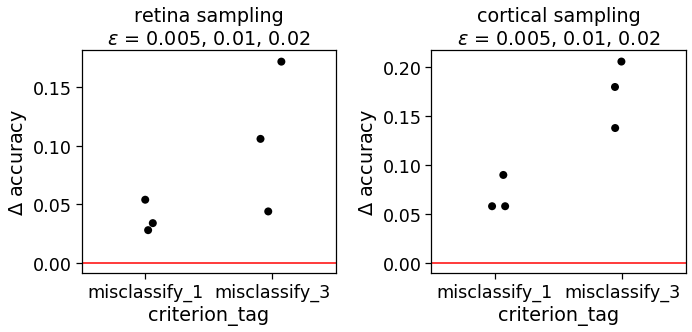}
\caption{Improvements in robustness for retinal and cortical fixations models from the best baseline for various adversarial criteria at small perturbations on ImageNet10.}
\label{fig:attack_confidentmistakes}
\end{figure}

We compared the robustness for $L_{\infty}$ PGD run for 5 iterations with a step size of $\epsilon$ /3 and the adversarial criteria of untargeted misclassified out of top-1 or top-3 predicted classes by model (misclassify\_1, misclassify\_3) and targeted misclassification (y\_adv = y\_true + 1) with adversarial class in top-1 or predicted with at least 80\% probability (targeted\_50, targeted\_80). Improvement in robustness of retinal and cortical fixations models was greater for misclassify\_3 compared to misclassify\_1. Improvement in robustness of retinal and cortical fixations models was approximately the same for targeted\_50 and targeted\_80 (see Figure \ref{fig:attack_confidentmistakes}).

\subsection{Quality of Adversarial Perturbations}

We explored how semantically meaningful (``quality'') large adversarial perturbations were for cortical fixations and retinal fixations models (see Figure \ref{fig:advexamples}). The large adversarial perturbations for the models were not anymore meaningful than for the coarse fixations model.

\subsection{Visibility of Adversarial Perturbations}

\begin{figure*}\centering
\includegraphics[width=\textwidth]{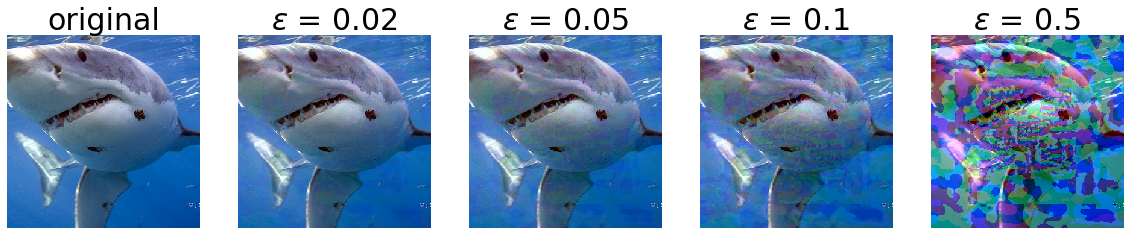}
\includegraphics[width=\textwidth]{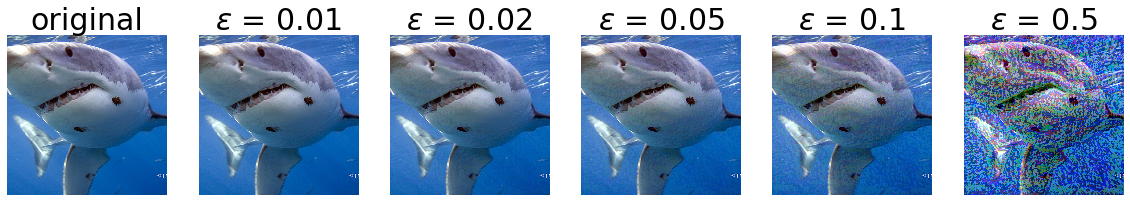}
\includegraphics[width=\textwidth]{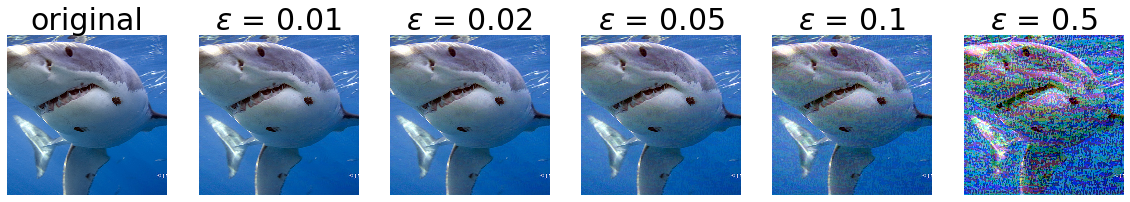}
\caption{Original image and adversarial examples at various $L_{\infty}$ distances $\epsilon$. Adversarial examples were generated for the cortical fixations model (top row), retinal fixations model (middle row), coarse fixations model (bottom row) using 5-step PGD with step size $\epsilon$ /3.}
\label{fig:advexamples}
\end{figure*}

We explored the $L_{\infty}$ distance at which the adversarial perturbations become visible to the naked eye when the adversarial example and the original image are presented side by side. We present the visualized scan for an image of a shark here (see Figure \ref{fig:advexamples}). At preliminary inspection, the adversarial perturbations seem to become visible around $\epsilon \approx$ 0.2 or 0.5.

\subsection{Training With Only Bounding Boxes}

\begin{figure}\centering
\includegraphics[width=\columnwidth]{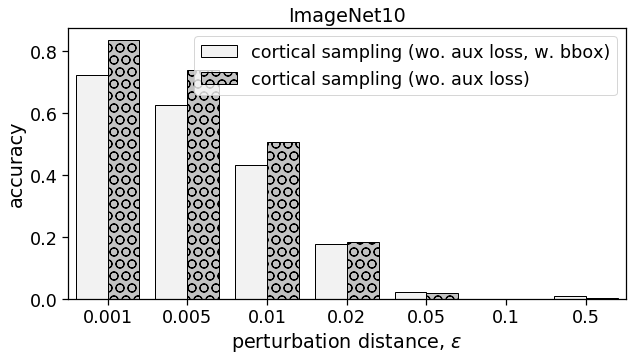}
\caption{Robustness of cortical fixations models trained on only the bounding boxes of training images ('w. bbox') and on the standard training images. Both models were trained without the standard auxiliary loss during training. Both models were evaluated on ImageNet10 to targeted (y\_adv = y\_true + 1) $L_{\infty}$ PGD attacks run for 5 iterations with step size $\epsilon$/3.}
\label{fig:bbox_train}
\end{figure}

We tried training the cortical fixations model on just the training images bounding boxes. We trimmed images in the training set of ImageNet10 (only training images, test images were left as is) to just their first bounding box and trained a cortical fixations model without the standard auxiliary loss during training. We compared this to a cortical fixations model trained on the usual training images without the standard auxiliary loss during training (see Figure \ref{fig:bbox_train}). 

Since only some of the images in the training set come with a bounding box annotation and since we only kept the first bounding box annotation in images, the model trained on the bounding boxes trained on far fewer samples. The model trained only on the bounding boxes did not show improved robustness compared to the model trained on the standard images.

\subsection{Alternate Cortical Sampling Implementations}

\begin{figure}\centering
\includegraphics[width=\columnwidth]{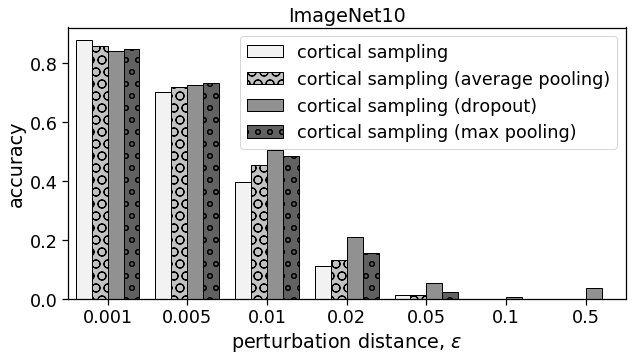}
\caption{Robustness of alternate cortical fixations models as compared to the cortical fixations model as used in the rest of the paper on ImageNet10 to targeted (y\_adv = y\_true + 1) $L_{\infty}$ PGD attacks run for 5 iterations with step size $\epsilon$/3.}
\label{fig:alt_mechanisms}
\end{figure}

We carried out some experiments with alternate methods of incorporating information across the scales (that result from a single fixation point) for the cortical sampling (see Figure \ref{fig:alt_mechanisms}). We experimented with max and average pooling across the scales and employing a large dropout during training on the concatenated representation.

For the cortical fixations models as used in the rest of the paper, we simply concatenated the representation from each scale to form the complete representation for each fixation. For the pooling alternatives, we pool the representations across the scales. For the large dropout alternative, we randomly drop 75\% of the elements in the final concatenated representation during training time. These alternate methods of incorporating information across the scales all performed better than the cortical sampling as used in the rest of the paper. This suggests that there are more robustness improvements to be gained from the cortical sampling models. 

\subsection{Combining Retinal And Cortical Sampling}

\begin{figure}\centering
\includegraphics[width=\columnwidth]{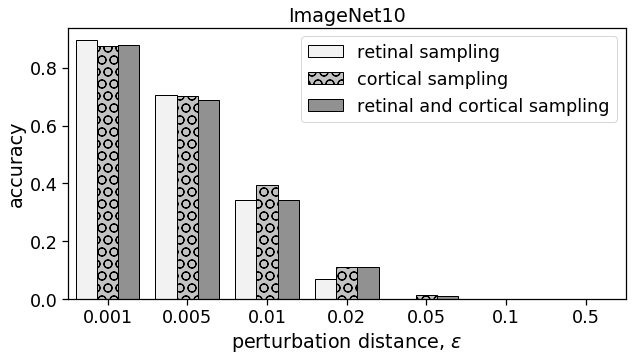}
\caption{Robustness of the retinal fixations models as compared to the cortical fixations model and the combined (retinal and cortical) fixations model on ImageNet10 to targeted (y\_adv = y\_true + 1) $L_{\infty}$ PGD attacks run for 5 iterations with step size $\epsilon$/3.}
\label{fig:combine_mechanisms}
\end{figure}

We attempted to combine the two proposed mechanisms, by first performing the retinal sampling on the image, then the cortical sampling. We envisioned this to be similar to how when presented with a visual stimulus, the re-sampling by the photoreceptors would occur before any further processing in the visual stream. Naively combining the two mechanisms in this manner resulted in an improvement in robustness from the baselines but the combined mechanisms model was only as robust as or less robust than either of the proposed mechanisms (see Figure \ref{fig:combine_mechanisms}).

\subsection{L2 CarliniWagner Attack}

We also conducted some preliminary experiments comparing the robustness of the models to the L2 variant of the CarliniWagner attack \cite{CWPGD} (see Table \ref{PGDCW_results}). We ran the attack for 100 iterations with the initial constant set to 0.01 for the ImageNet10 models. The preliminary results are consistent with our reported results with PGD (proposed mechanisms outperform the baselines).

\begin{table}[t]
\caption{Accuracy of the models under $L_2$ CarliniWagner attack with the initial constant set to 0.01 on ImageNet10. Retinal and cortical fixations models outperformed the baselines (standard ResNet, coarse fixations).}
\label{PGDCW_results}
\vskip 0.15in
\begin{center}
\begin{small}
\begin{sc}
\begin{tabular}{lccr}
\toprule
             mechanism &  accuracy \\
\midrule
  vanilla ResNet &  0.692 \\
    coarse fixations &  0.71 \\
 retinal fixations &  0.81 \\
 cortical fixations &  0.758 \\
\bottomrule
\end{tabular}
\end{sc}
\end{small}
\end{center}
\vskip -0.1in
\end{table}

\section{Gradient Obfuscation}

\subsection{Iterative Attacks Outperform Single-Step Attacks}

To check that gradient obfuscation was not an issue with the models, we verified that PGD was always more successful or as successful as FGSM when generating adversarial examples (see Table \ref{PGD_vs_FGSM}). We verified this with $L_{\infty}$ PGD run for 5 iterations against the equivalent FGSM, with the adversarial criteria that the true class is not in the top-3 classes predicted by the model.

\begin{table}[t]
\caption{Comparison of accuracy under PGD and equivalent FGSM attack. Top Table: Retinal Fixations Model. Bottom Table: Cortical Fixations Model. PGD consistently performs as well as or outperforms FGSM.}
\label{PGD_vs_FGSM}
\vskip 0.15in
\begin{center}
\begin{small}
\begin{sc}

\begin{tabular}{lrr}
\toprule
$\epsilon$ &  accuracy (FGSM) & accuracy (PGD) \\
\midrule
0.500 &                 0.054 & 0.000\\
0.100 &                 0.240 & 0.000  \\
0.050 &                 0.378 & 0.014  \\
0.020 &                 0.540 & 0.236 \\
0.010 &                 0.698 & 0.600 \\
0.005 &                 0.850 & 0.840 \\
0.001 &                 0.900 & 0.900 \\
\bottomrule

\toprule
$\epsilon $&  accuracy (FGSM) & accuracy (PGD) \\
\midrule
0.500 &                 0.108 & 0.010 \\
0.100 &                 0.326 & 0.036\\
0.050 &                 0.440 & 0.098\\
0.020 &                 0.636 & 0.344\\
0.010 &                 0.742 & 0.638\\
0.005 &                 0.834 & 0.818\\
0.001 &                 0.886 & 0.886 \\
\bottomrule

\end{tabular}

\end{sc}
\end{small}
\end{center}
\vskip -0.1in
\end{table}

We saw that PGD consistently outperformed or at the very least performed as well as FGSM, this suggests that following the direction of gradient descent is useful for constructing adversarial examples and that gradient obfuscation is likely not an issue with the models.

\subsection{White-Box Attacks Outperform Black-Box}

If gradient obfuscation is not an issue with the models, white-box attacks (attacks that directly use the model gradients) should always outperform black-box attacks.

\subsubsection{Transfer Attack}

We performed a transfer attack from the standard ResNet model to the retinal fixations and cortical fixations models (see Table \ref{transferattack_table}). We compared the accuracy of the models under this transfer attack and accuracy under standard PGD. We perform $L_{\infty}$ PGD for 5 steps, with a step size of $\epsilon$/3, maximum adversarial perturbation $\epsilon$ = 0.005 or $\epsilon$ = 0.5. We perform targeted attacks with desired adversarial class ($y+1 \mod n$, $n=10$) and desired predicted probability of at least $0.8$.

For each perturbation limit, we first run PGD on the images in the ImageNet10 test split that were correctly predicted by the standard ResNet model. We then evaluate the retinal fixations and cortical fixations models on these adversarial examples generated for the standard ResNet model (tarnsfer attack). For the comparison to an equivalent white-box attack, we then run PGD on the retinal fixations and cortical fixations models for the images in ImageNet10 test split that were correctly predicted by both the standard ResNet model and the retinal or cortical fixations model.

\begin{table}[t]
\caption{Accuracy of the retinal fixations and cortical fixations models was lower under a standard PGD attack performed directly on the models compared to a transfer attack from a standard ResNet model.}
\label{transferattack_table}
\vskip 0.15in
\begin{center}
\begin{small}
\begin{sc}
\begin{tabular}{lcccr}
\toprule
            mechanism &    attack method &  accuracy \\
\midrule
 retinal &              PGD ($\epsilon$ = 0.005) &              0.560694 \\
 retinal &  transfer ($\epsilon$ = 0.005) &              0.933333 \\
    cortical&              PGD ($\epsilon$ = 0.005) &              0.549133 \\
    cortical  &  transfer ($\epsilon$ = 0.005) &              0.887179 \\
 retinal &              PGD ($\epsilon$ = 0.5) &              0.000000 \\
 retinal &  transfer ($\epsilon$ = 0.5) &              0.181208 \\
    cortical &              PGD ($\epsilon$ = 0.5) &              0.000000 \\
    cortical &  transfer ($\epsilon$ = 0.5) &              0.261745 \\
\bottomrule
\end{tabular}
\end{sc}
\end{small}
\end{center}
\vskip -0.1in
\end{table}

The transfer attacks consistently underperformed at generating adversarial examples compared to a direct white-box PGD attack. This suggests that gradient obfuscation is not an issue with the models.

\subsubsection{Boundary Attack}

We compared the adversarial examples generated with boundary attack (black-box decision-based attack, \cite{BoundaryAttack}) with PGD (white-box gradient-based attack) (see Table \ref{boundaryattack_table}).

Both attacks were run in a targeted setting on models trained on ImageNet10. We ran it on a single image of a panda (ILSVRC2012\_val\_00042615.JPEG) to be misclassified as a shark. We initialized the boundary attack with an image of a shark (ILSVRC2012\_val\_00029481.JPEG). We only ran the comparison for a single image because boundary attack takes a while to run. We ran boundary attack for 1000 iterations and PGD for 5 iterations, comparing the size of adversarial perturbations.

\begin{table}[t]
\caption{Size of adversarial perturbations generated with 5 steps of PGD was always less than with 1000 steps of boundary attack. Here, we refer to the standard ResNet model as 'vanilla', coarse fixations as 'coarse', retinal fixations as 'retinal', cortical fixations as 'cortical'.}
\label{boundaryattack_table}
\vskip 0.15in
\begin{center}
\begin{small}
\begin{sc}
\begin{tabular}{lcccr}
\toprule
            mechanism &    attack method &  distance \\
\midrule
 cortical &              PGD &              8.2e-06 \\
 cortical &              boundary attack &              2.5e-03 \\
 retinal &              PGD &              1.0e-05 \\
 retinal &              boundary attack &              1.4e-03 \\
 coarse  &              boundary attack &              1.1e-03 \\
 vanilla &              boundary attack &              6.45e-04 \\
\bottomrule
\end{tabular}
\end{sc}
\end{small}
\end{center}
\vskip -0.1in
\end{table}

Generated adversarial perturbations were always smaller (better) with 5 steps of PGD compared to 1000 steps of boundary attack. This again suggests that gradient obfuscation is not an issue with the models. The size of adversarial perturbations for the retinal and cortical fixations models generated with boundary attack was also larger than for the coarse fixations or standard ResNet model.

The probability assigned to the mispredicted class for examples generated with 5 steps PGD was also much greater than for 1000 steps of boundary attack ($\approx$100\% for PGD vs $\approx$50\% for boundary attack). This is however an unfair comparison since PGD explicitly optimizes for the probability of the mispredicted class while boundary attack does not, but we mention it here for completeness.

\subsubsection{Backward Pass Differentiable Approximation}

We also attempted to generate adversarial examples using the backward pass differentiable approach (BDPA) \cite{GradientObfuscation} (see Table \ref{BPDA_table}). If gradient obfuscation is not an issue with the models, this surrogate gradient approach should not outperform directly using the full model gradient with PGD.

For the retinal fixations model, we approximated the retina transform $g(x)$ as $g(x) \approx x$ for the gradient (backward pass). The cortical fixations model performs several transforms $h(i(j(x)))$ for each scale fragment. We approximated $i(x)$ as $i(x) \approx x$ for the gradient (backward pass). We did not approximate the gradients for the subsample and cropping operation further. We performed $L_{\infty}$ PGD on one image of a panda (top in Table $\ref{BPDA_table}$) and one image of a frog (bottom in Table $\ref{BPDA_table}$).

\begin{table}[t]
\caption{Ability to find an adversarial example with standard PGD was always superior to the BPDA PGD. Top, Bottom Tables: Ability to find adversarial examples using various hyperparameters with two different pictures as the original image.}
\label{BPDA_table}
\vskip 0.15in
\begin{center}
\begin{small}
\begin{sc}
\begin{tabular}{lcccr}
\toprule
            mechanism &    attack method (steps, $\epsilon$) & adv found \\
\midrule
 cortical &              PGD (5,0.005) &              Y \\
 cortical &              PGD BPDA (60,0.005) &       N  \\
  cortical &              PGD BPDA (5,0.5) &       Y  \\
 retinal &              PGD (5,0.005) &              Y \\
 retinal &             PGD BPDA (120,0.005) &              N \\
  retinal &             PGD BPDA (5,0.5) &              Y \\
\bottomrule

\toprule
            mechanism &    attack method (steps, $\epsilon$) & adv found \\
\midrule
 cortical &              PGD (5,0.01) &              Y \\
 cortical &              PGD BPDA (300,0.01) &       N  \\
  cortical &              PGD BPDA (5,0.5) &       Y  \\
 retinal &              PGD (10,0.005) &              Y \\
 retinal &             PGD BPDA (300,0.005) &              N \\
  retinal &             PGD BPDA (10,0.5) &              Y \\
\bottomrule
\end{tabular}
\end{sc}
\end{small}
\end{center}
\vskip -0.1in
\end{table}

Standard PGD, taking the gradient through the full model, was always superior to PGD BPDA at finding an adversarial example. This suggests that gradient obfuscation is not an issue with the models.

We also verified that PGD BDPA was able to find adversarial examples when we allow very large perturbations.


\section{Image Sources}

The original pictures of the peacock (e.g. see Figure 1: Bottom Right in main paper) and checkerboard (e.g. seeFigure 1: Bottom Left in main paper) used to visualize the effect of the biologically inspired mechanisms were obtained from online sources. The image of a peacock was taken from Flickr (see \url{www.flickr.com/photos/kkoshy/32401990166}). The author, Koshy Koshy (see \url{https://www.flickr.com/photos/kkoshy/}), explicitly authorized the image for reuse and modification under the Attribution 2.0 Generic (CC BY 2.0) license (\url{https://creativecommons.org/licenses/by/2.0/}). The image of a checkerboard was taken from Wikimedia Commons (see \url{https://en.m.wikipedia.org/wiki/File:Checkerboard_reflection.svg}). The author, M. W. Toews (see \url{https://commons.wikimedia.org/wiki/User:Mwtoews}), explicitly authorized the image for reuse and modification under the Attribution-ShareAlike 4.0 International (CC BY-SA 4.0) license (\url{https://creativecommons.org/licenses/by-sa/4.0/deed.en}).

\end{document}